\documentclass[11pt,a4paper]{article}
\usepackage[hyperref]{emnlp-ijcnlp-2019}
\usepackage{times}
\usepackage{latexsym}

\usepackage{url}
\usepackage{stfloats}
\usepackage{url}
\usepackage{float}
\usepackage{graphicx}
\usepackage{epstopdf}
\usepackage{amsfonts}
\usepackage{amssymb}
\usepackage{amsmath}
\usepackage{verbatim}
\usepackage{multirow}
\usepackage{amsmath}
\usepackage[noend]{algorithmic}
\usepackage[ruled]{algorithm2e}
\usepackage{subfigure}
\usepackage{pifont}
\usepackage{array}
\usepackage{fp}
\usepackage{makecell}
\usepackage{flushend}
\usepackage{cases}
\usepackage{color}
\usepackage{booktabs}
\usepackage{bbm}
\usepackage[ruled]{algorithm2e}
\usepackage{multirow}

\usepackage{CJK}
\usepackage{listings}
\usepackage{xcolor}
\usepackage{mathrsfs}

\newcommand{\ie}{\emph{i.e.,}\xspace}

\newcommand{\eg}{\emph{e.g.,}\xspace}
\newcommand{\ignore}[1]{}

\aclfinalcopy 

\title{CRSLab: An Open-Source Toolkit for Building Conversational Recommender System}

\author{
\setcounter{footnote}{1}
	Kun Zhou\textsuperscript{\rm{1},\rm{4}} \thanks{$^\dagger$ Equal contribution.}, 
	Xiaolei Wang\textsuperscript{\rm{2}} $^\dagger$,
	Yuanhang Zhou\textsuperscript{\rm{1},\rm{4}},
	Chenzhan Shang\textsuperscript{\rm{2}},
	Yuan Cheng\textsuperscript{\rm{3}}, \\
\setcounter{footnote}{0}
	\textbf{Wayne Xin Zhao}\textsuperscript{\rm{2},\rm{4}} \thanks{$^*$ Corresponding author, Email: batmanfly@gmail.com.},
	\textbf{Yaliang Li}\textsuperscript{\rm{5}}, \and
	\textbf{Ji-Rong Wen}\textsuperscript{\rm{1},\rm{2},\rm{4}} \\
	
	\textsuperscript{1}School of Information, Renmin University of China \\
	\textsuperscript{2}Gaoling School of Artificial Intelligence, Renmin University of China \\
	\textsuperscript{3}School of Statistics, Renmin University of China \\
	\textsuperscript{4}Beijing Key Laboratory of Big Data Management and Analysis Methods \\
	\textsuperscript{5}Alibaba Group \\
}

\hypersetup{draft}
\begin{document}
\maketitle
\begin{abstract}
In recent years, conversational recommender system (CRS) has received much attention in the research community.  However, existing studies on CRS vary in scenarios, goals and techniques, lacking unified, standardized implementation or comparison. To tackle this challenge, we propose an open-source CRS toolkit CRSLab, which provides a unified and extensible framework with highly-decoupled modules to develop CRSs. Based on this framework, we collect 6 commonly-used human-annotated CRS datasets and implement 18 models that include recent techniques such as graph neural network and pre-training models. Besides, our toolkit provides a series of automatic evaluation protocols and a human-machine interaction interface to test and compare different CRS methods. The project and documents are released at 
\textcolor{blue}{\url{https://github.com/RUCAIBox/CRSLab}}.
\end{abstract}

\section{Introduction}
Recent years have witnessed remarkable progress in the conversational recommender system (CRS)~\cite{DBLP:conf/kdd/Christakopoulou16,DBLP:conf/sigir/SunZ18,DBLP:conf/nips/LiKSMCP18}, which aims to provide high-quality recommendations to users through natural language conversations.
To build an effective CRS, users have proposed a surge of datasets~\cite{DBLP:conf/emnlp/KangBSCBW19,DBLP:conf/coling/ZhouZZWW20,DBLP:conf/acl/LiuWNWCL20} and models~\cite{DBLP:conf/wsdm/Lei0MWHKC20,DBLP:conf/emnlp/ChenLZDCYT19,DBLP:journals/corr/abs-1907-00710}.
However, these works are different in scenarios (\eg movies or E-commerce platform), goals (\eg accurate recommendation or user activation) and techniques (\eg graph neural network or pre-training models),
hence it is challenging for users to quickly set up reasonable baseline systems or develop new CRS models.

To alleviate the above issues, we have developed \textbf{CRSLab}, the first open-source CRS toolkit for research purpose.
In CRSLab, we offer a unified and extensible framework with highly-decoupled modules to develop a CRS.
Specifically, we unify the task description of existing works for CRS into three sub-tasks, namely \emph{recommendation}, \emph{conversation} and \emph{policy}, covering the common functional requirements of mainstream CRSs.
To implement the overall framework, we design and develop highly-decoupled modules (\eg data modules and model modules), which provide clear interfaces.
Besides, we encapsulate useful procedures and common functions shared by different modules for reuse.
In this way, it is easy for users to add new datasets or develop new models with our toolkit. 

Based on the framework, we integrate comprehensive benchmark datasets and models in CRSLab. So far, we have incorporated 6 commonly-used human-annotated datasets and implemented 18 models, including advanced techniques such as graph neural network (GNN)~\cite{DBLP:conf/esws/SchlichtkrullKB18,DBLP:conf/kdd/ZhouZBZWY20} and pre-training models~\cite{DBLP:conf/naacl/DevlinCLT19,DBLP:conf/cikm/ZhouZWWZWW20}.
To support these models, we perform necessary preprocessing on integrated datasets (\eg entity linking and word segmentation), and release the processed data.
In our CRSLab, we provide flexible supporting mechanisms via the configuration files or command lines to run, compare and test these models on integrated datasets, by which users can develop a powerful CRS.

Furthermore, CRSLab provides a series of automatic evaluation protocols and a human-machine interaction interface for testing and comparing different CRSs, which are useful to standardize the evaluation protocol for conversational recommendation. 
Specifically, we implement various automatic evaluation metrics to test a CRS on recommendation, conversation and policy tasks, respectively, covering commonly-used metrics in existing works.
In addition, CRSLab provides a human-machine interactive interface to perform quantitative analysis, which is helpful for users to deploy their systems and converse with the systems via the webpage.

\section{Background}
\label{sec-problem}
As aforementioned, existing CRS datasets and models vary in scenarios and goals, thus the corresponding domains and task definitions can be different, which brings gaps when applying existing models on different datasets or scenarios.
To fill these gaps, based on previous works~\cite{DBLP:conf/wsdm/Lei0MWHKC20,DBLP:conf/coling/ZhouZZWW20,DBLP:conf/sigir/SunZ18}, we unify the task of CRS into two basic sub-tasks and an auxiliary sub-task, namely recommendation, conversation and policy. 
These three sub-tasks are described as: given the dialog context (\ie historical utterances) and other useful side information (\eg interaction history and knowledge graph), we aim to (1) predict user-preferred items (recommendation), (2) generate a proper response (conversation), and (3) select proper interactive action (policy).

It is worth noting that the above task description covers most of CRS models and datasets.
The recommendation and the conversation sub-tasks have been considered by all of these works. 
The policy sub-task is needed by recent works~\cite{DBLP:conf/coling/ZhouZZWW20,DBLP:conf/wsdm/Lei0MWHKC20}, by which the CRS can proactively guide the dialog for better recommendation.
For different goals and scenarios, the policy sub-task can be different.
For example, TG-ReDial~\cite{DBLP:conf/coling/ZhouZZWW20} utilizes a topic prediction model to accomplish the policy sub-task, while DuRecDial~\cite{DBLP:conf/acl/LiuWNWCL20} defines it as a goal planning task.

\section{CRSLab}

\begin{figure}[t]
\centering
\includegraphics[width=\linewidth]{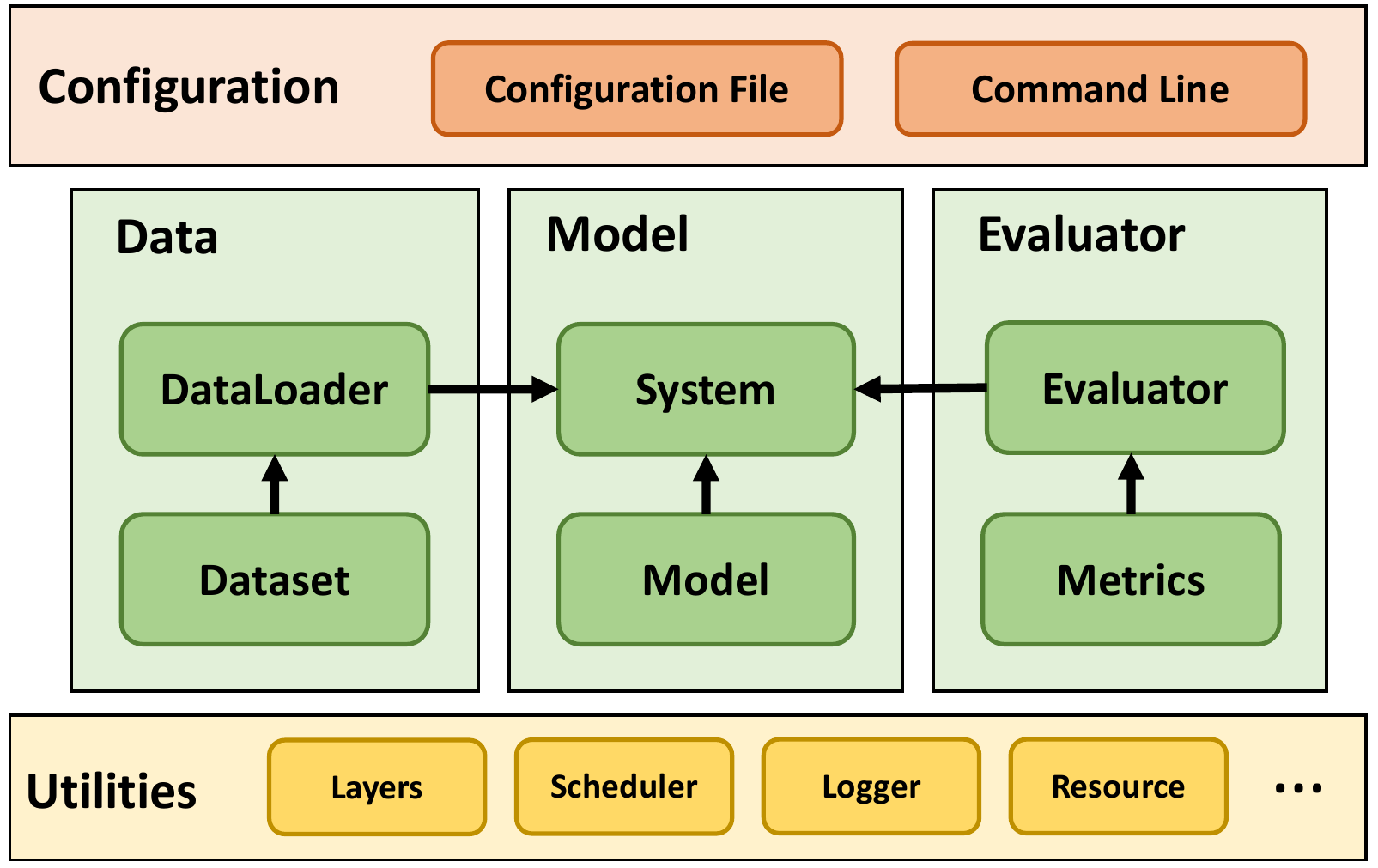}
\caption{The overall framework of CRSLab.}
\label{fig-frame}
\end{figure}

The overall framework of our toolkit CRSLab is presented in Figure~\ref{fig-frame}. 
The configuration module provides a flexible interface for users to easily set up the experiment environment (\eg datasets, models and hyperparameters).
The data, model and evaluation modules are built upon the configuration module, which forms the core part of our toolkit.
The bottom part is the utility module, providing auxiliary functions and interfaces for reuse in other modules (\eg logger and resource).
In the following part, we briefly present the designs of the above modules, and more details can be found in the toolkit documents.

\subsection{Configurations}
In CRSLab, we design the configuration module for users to conveniently select or modify the experiment setup (\eg dataset, model and hyperparameters).
Specifically, we design the class $\mathsf{Config}$ to store all the configuration settings, which specifies the model and its hyperparameters for each component of the CRS and environment for a given experiment.
To avoid specifying complicated command line parameters, we provide a few commonly-used configured settings (\ie file path and debug mode) in the command line while others in YAML configuration files. 
In this way, users can build and evaluate a variety of different CRSs with only slight modifications in the configuration files.

\subsection{Data Modules}
For extensibility and reusability, we design an elegant data flow that transforms raw dataset into the model input as following: Raw Public Dataset $\longrightarrow$ Preprocessed Dataset $\longrightarrow$ $\mathsf{Dataset}$ $\longrightarrow$ $\mathsf{DataLoader}$ $\longrightarrow$ $\mathsf{System}$.
Next, we detail the design of these components.

\subsubsection{Data Preprocessing}
\begin{table*}[t]
\centering
\small
\begin{tabular}{c|c|c|c|c|c|c}
\toprule
Dataset &Dialog &Utterance &Domain &Policy Model &Entity KG &Word KG \\ \midrule
ReDial~\cite{DBLP:conf/nips/LiKSMCP18} &10,006 &182,150 &Movie &-- &DB &CNet \\ 
TG-ReDial~\cite{DBLP:conf/coling/ZhouZZWW20} &10,000 &129,392 &Movie &Topic Prediction  &CN-DB &HNet \\ 
GoRecDial~\cite{DBLP:conf/emnlp/KangBSCBW19} &9,125 &170,904 &Movie &Action Prediction &DB &CNet \\	 		
DuRecDial~\cite{DBLP:conf/acl/LiuWNWCL20} &10,200 &156,000 &Movie, Music &Goal Planning &CN-DB &HNet \\ 
INSPIRED~\cite{DBLP:conf/emnlp/HayatiKZSY20} &1,001 &35,811 &Movie & Strategy Prediction &DB &CNet \\
OpenDialKG~\cite{DBLP:conf/acl/MoonSKS19} &13,802 &91,209 &Movie, Book &Path Generation &DB &CNet \\
\bottomrule
\end{tabular}
\caption{The collected datasets in CRSLab. DB and CN-DB stand for the entity-oriented knowledge graph DBpedia and CN-DBpedia, respectively. CNet and HNet stand for the word-oriented knowledge graph ConceptNet and HowNet, respectively.}
\label{table:dataset}
\end{table*}

Since raw public datasets vary in formats and features, we preprocess these datasets to support unified interfaces in data modules.
Based on the task description in Section~\ref{sec-problem}, we first preprocess CRS datasets to match the input and output formats.
Specifically, we organize the dialog context and side information as the input while extract the recommended items, dialog actions and responses as the output of recommendation, policy and conversation sub-tasks, respectively.
To support some advanced models (\eg graph neural network and pre-training models), we incorporate useful side data (\eg knowledge graph) and conduct specific preprocessing (\eg entity link and BPE segment).

As shown in Table~\ref{table:dataset}, we have collected 6 commonly-used human-annotated datasets and released the preprocessed versions with the side data in our CRSLab.
Besides, we also release the pre-trained word embeddings and other associated files, which ease the use of integrated datasets and reduce the time cost.

\subsubsection{Dataset Class}
To decouple the implementation of data preparing in CRSLab, we design the class $\mathsf{Dataset}$ for integrating the model-independent data processing functions, while the rest functions are implemented by the class $\mathsf{DataLoader}$.
In this way, $\mathsf{Dataset}$ only focuses on processing the input data into a unified format (\ie a list of $\mathsf{python.dict}$), without considering specific models.
In CRSLab, we design the class $\mathsf{BaseDataset}$ which includes some common attributes (\eg configurations and data path) and basic functions (\eg load data) of $\mathsf{Dataset}$, hence users can inherit $\mathsf{BaseDataset}$ with very few modifications to integrate new datasets.

\subsubsection{DataLoader Class}
Indeed, different CRS models need different formats.
Since $\mathsf{Dataset}$ has processed the input data into a unified format, $\mathsf{DataLoader}$ further reformulates data for supporting various models.
Specifically, $\mathsf{DataLoader}$ focuses on selecting features from the processed data after $\mathsf{Dataset}$ to form tensor data(\ie $\mathsf{torch.Tensor}$) in a batch or mini-batch, which can be directly used for the update and computation of downstream models.
To implement it, we design the class $\mathsf{BaseDataLoader}$ to integrate common attributes and functions, and inherit it to produce new dataloaders for corresponding models.

\subsection{Model Modules}
Based on the task description and above data modules, we reorganize the implementations of existing CRS in a hierarchical framework, 
in which the model module provides functions and interfaces for building and running specific models, while the system module trains or evaluates contained models for accomplishing the defined task.

\subsubsection{Model Class}
\begin{table*}[t]
\centering
\small
\begin{tabular}{l|c|c|c|c}
\toprule
Category &Model &GNN &PTM &Reference\\ \midrule
\multirow{4} * {CRS model} &ReDial &$\times$ &$\times$ &\cite{DBLP:conf/nips/LiKSMCP18} \\
 &KBRD &$\surd$ &$\times$ &\cite{DBLP:conf/emnlp/ChenLZDCYT19}\\ 
 &KGSF &$\surd$ &$\times$ &\cite{DBLP:conf/kdd/ZhouZBZWY20}\\ 
 &TG-ReDial &$\times$ &$\surd$ &\cite{DBLP:conf/coling/ZhouZZWW20}\\ \midrule
\multirow{6} * {Recommendation model} &Popularity &$\times$ &$\times$ &-- \\ 
 &GRU4Rec &$\times$ &$\times$ &\cite{DBLP:journals/corr/HidasiKBT15}\\ 
 &SASRec &$\times$ &$\times$ &\cite{DBLP:conf/icdm/KangM18}\\ 
 &TextCNN &$\times$ &$\times$ &\cite{DBLP:conf/emnlp/Kim14}\\
 &R-GCN &$\surd$ &$\times$ &\cite{DBLP:conf/esws/SchlichtkrullKB18} \\ 
 &BERT &$\times$ &$\surd$ &\cite{DBLP:conf/naacl/DevlinCLT19}\\ \midrule
\multirow{3} * {Conversation model} &HERD &$\times$ &$\times$ &\cite{DBLP:conf/aaai/SerbanSBCP16} \\ 
 &Transformer &$\times$ &$\times$ &\cite{DBLP:conf/nips/VaswaniSPUJGKP17} \\ 
 &GPT-2 &$\times$ &$\surd$ &\cite{radford2019language}\\ \midrule
\multirow{5} * {Policy model} &PMI &$\times$ &$\times$ &--\\
 &MGCG &$\times$ &$\times$ &\cite{DBLP:conf/acl/LiuWNWCL20}\\ 
 &Conv-BERT &$\times$ &$\surd$ &\cite{DBLP:conf/coling/ZhouZZWW20}\\
 &Topic-BERT &$\times$ &$\surd$ &\cite{DBLP:conf/coling/ZhouZZWW20} \\
 &Profile-BERT &$\times$ &$\surd$ &\cite{DBLP:conf/coling/ZhouZZWW20}\\ 
\bottomrule
\end{tabular}
\caption{The implemented models in CRSLab. Recommendation, policy and conversation models specify corresponding individual sub-task, while CRS models can accomplish these sub-tasks together. GNN and PTM stand for the graph neural network and pre-training models, respectively.}
\label{table:model}
\end{table*}

As mentioned before, a CRS may consist of several models for corresponding sub-tasks. In the model module, we focus on providing a basic structure and useful highly-decoupled functions or procedures for development.
Specifically, we unify the basic attributes and functions of various models (\eg parameter initialization and model loading) into the class $\mathsf{BaseModel}$. A user can inherit $\mathsf{BaseModel}$ and implement a few functions to develop and design new models.

We have carefully surveyed the recent literature and selected commonly-used models in four categories, namely CRS models, recommendation models, conversation models and policy models.
Among them, CRS models integrate the recommendation model and the conversation model to improve both models, while recommendation, policy and conversation models only focus on one individual sub-task.
As illustrated in Table~\ref{table:model}, we mainly focus on recently proposed neural methods, and also keep some classic heuristic methods such as Popularity and PMI.
In the first release version, we have implemented 18 models, including some advanced models such as graph neural networks and pre-training models.
For all the implemented models, we have tested their performance on two or three selected datasets, and invited a code reviewer to examine the correctness of the implementation.
In the future, more methods will also be incorporated along with regular updates.

\subsubsection{System Class}
To support flexible architectures for CRS at a high level, we devise the system module which serves as a junction to integrate the dataloader, model and evaluator modules for building a complete CRS.
Specifically, the system module mainly aims to set up models for accomplishing the CRS task, distribute the tensor data from dataloader to corresponding models, train the models with proper optimization strategy, and conduct evaluation with specified protocols.

To implement the above requirements, we design the class $\mathsf{BaseSystem}$ to unify the structure and interfaces, which contains corresponding functions. In $\mathsf{BaseSystem}$, we also implement a series of useful functions, such as optimizer initialization, learning ratio adjustment and early stop strategy. These functions and tiny tricks ease the developing process of new system and largely improve the user experiences with our CRSLab.

\subsection{Evaluation Modules}
The function of the evaluation module is to implement the evaluation protocols for CRS models.
In CRSLab, we implement commonly-used automatic evaluation metrics.
Besides we also design a human-machine interactive interface for users to perform an end-to-end quantitative analysis.

\subsubsection{Automatic Evaluation Modules}
\begin{table*}[t]
\small
\centering
\begin{tabular}{l|p{9cm}}
\toprule
Category &Metrics \\ \midrule
Recommendation Metrics &Hit@\{1,10,50\}, MRR@\{1,10,50\}, NDCG@\{1,10,50\} \\ \midrule
Conversation Metrics &Perplexity, BLEU-\{1,2,3,4\}, Embedding Average/Extreme/Greedy, Distinct-\{1,2,3,4\}  \\ \midrule
Policy Metrics &Accuracy, Hit@\{1,3,5\} \\
\bottomrule
\end{tabular}
\caption{The implemented automatic evaluation metrics in CRSLab.}
\label{table:metrics}
\end{table*}

Since the CRS task is divided into three sub-tasks, we develop corresponding automatic metrics in the evaluation module. We summarize all the supported automatic evaluation metrics in Table~\ref{table:metrics}.
For recommendation sub-task, following existing CRS models~\cite{DBLP:conf/sigir/SunZ18,DBLP:conf/cikm/ZhangCA0C18}, we develop ranking-based metrics for measuring the ranking performance of the generated recommendation lists by a CRS.
For conversation sub-task, CRSLab supports both relevance-based and diversity-based evaluation metrics.
The relevance-based metrics include Perplexity, BLEU~\cite{DBLP:conf/acl/PapineniRWZ02} and Embedding metrics~\cite{DBLP:conf/emnlp/LiuLSNCP16}, which measures the similarity between ground-truth and generated responses from the perspective of probability, n-gram and word embedding, respectively.
The diversity-based metrics are Distinct-\{1,2,3,4\}~\cite{DBLP:conf/naacl/LiGBGD16}, measuring the number of distinct \{1,2,3,4\}-gram in the generated responses.
Since the policy sub-task varies in existing CRSs (\eg action and topic prediction), we implement the commonly-used metrics Accuracy and Hit@K to evaluate the performance between the true and predicted values.

Similarly, we design the class $\mathsf{BaseEvaluator}$ by implementing common attributes and functions.
Then, we inherit $\mathsf{BaseEvaluator}$ and implement $\mathsf{RecEvaluator}$, $\mathsf{ConvEvaluator}$ and $\mathsf{PolicyEvaluator}$ for evaluating recommendation, conversation and policy sub-tasks, respectively.
It is worth noting that we implement $\mathsf{report()}$ function in these evaluators. With this function users can print and monitor the performance of models evaluating on validation or test set.

\subsubsection{Human-Machine Interaction Interface}
To evaluate a CRS quantitatively, CRSLab offers a human-machine interaction interface to help users perform an end-to-end evaluation.
The human-machine interaction interface is integrated with the system module, by which the interaction strategy within the interface can be easily adapted for a specific policy model.
In this way, a user can converse with a CRS and diagnose the system, which provides an approach to directly evaluating the overall performance of a CRS.
Besides, the interaction interface enables users to correct errors by modifying intermediate results.

Specifically, to perform end-to-end evaluation, users first set up the background of a simulated user (\eg interaction history and user profile), then freely chat with the CRS through the interface.
During a conversation, the dialog history and the output of each component (including the recommended items and selected policy) are stored as a dictionary, which helps users get a good understanding of how their system works.

\subsection{Utilities}
In order to better use our CRSLab, 
we design the utility module which includes auxiliary functions (\eg $\mathsf{logger()}$ and $\mathsf{scheduler()}$).
Specifically, we implement a series of useful functions to facilitate the use of our toolkit. 
A particularly useful function is $\mathsf{scheduler()}$, which provides a set of strategies for training large-scale models, such as warming-up strategy and weight decay.
Besides, we also implement other functions to improve the user experiences with our toolkit, such as $\mathsf{save\_model()}$ and $\mathsf{load\_model()}$ to store and reuse the learned models, $\mathsf{logger()}$ to print and monitor the running process.

To ease the development of a new CRS, we also decouple commonly-used functions or procedures (\eg $\mathsf{Layers}$) in other modules to form the utility file (\ie utils.py), which constitutes another part of the utility module.
In this way, users can assemble or slightly modify functions in utility files to develop and design a new CRS.

\section{System Demonstration}
\begin{figure*}[t]
\centering
\includegraphics[width=\linewidth]{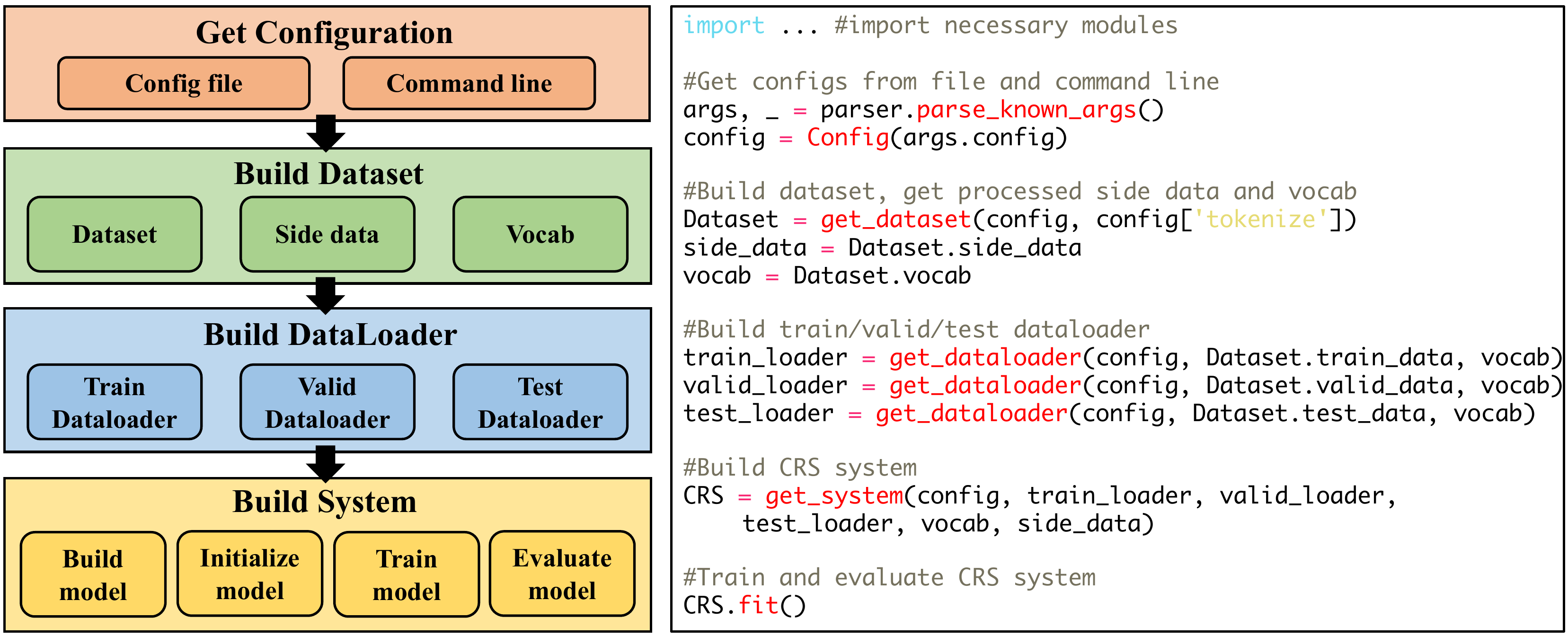}
\caption{An illustrative usage flow of our CRSLab.}
\label{fig-demo}
\end{figure*}

In this section, we show how to use our CRSLab with code examples. We detail the usage description in two parts, namely running an existing CRS in our toolkit and implementing a new CRS based on the interfaces provided in our toolkit.

\subsection{Running an Existing CRS}
Our CRSLab allows for easy creation of a CRS within a few lines of code. 
Figure~\ref{fig-demo} presents a general procedure for running an existing CRS in our toolkit.

To begin with, the whole procedure relies on the configuration to prepare the dataset and build the system.
In the configuration, the user selects a dataset to use and specifies the tokenizer.
Then, the $\mathsf{Dataset}$ class will automatically download the dataset and perform necessary processesing steps (\eg tokenize and convert tokens to IDs) based on the configurations.
This procedure is executed by the function $\mathsf{get\_dataset()}$.
Based on the processed datasets, users can use the function $\mathsf{get\_dataloader()}$ to generate training, validation and test sets, in which the configurations specify the batch size and other parameters for data processing.
After that, users can adopt the function $\mathsf{get\_system()}$ to build a CRS, which leverages the prepared side data from dataset and the above dataloaders.
In the CRS, the configurations specify the structure of models and set up the training and evaluation procedures. users can start the running process by the following function $\mathsf{System.fit()}$.

\subsection{Adding a New CRS}
Based on our toolkit, it is convenient to implement a new CRS with provided interfaces. The user only needs to inherit a few basic classes and implement some interface functions.
In this part, we will introduce the detailed implementation process of adding a new dataset and model, respectively.

\subsubsection{Adding a New Dataset}
To add a new dataset, one needs to inherit $\mathsf{BaseDataset}$ to design a new $\mathsf{Dataset}$ class for preparing the dataset into a unified format.
In $\mathsf{Dataset}$, the following functions are required to be implemented: $\mathsf{\_\_init\_\_()}$, $\mathsf{\_load\_data()}$ and $\mathsf{\_data\_preprocess()}$.

Specifically, in $\mathsf{\_\_init\_\_()}$, users set up parameters and the dataset links.
In $\mathsf{\_load\_data()}$, the training, validation, test data and other side data are loaded from corresponding files. Noting that if users follow our naming protocol, all they need to do is to reuse the implemented functions from the existing $\mathsf{Dataset}$ class.
The function $\mathsf{\_data\_preprocess()}$ performs the preparing of the loaded data.
We integrate useful functions in the utility module to ease the implementation.

\subsubsection{Adding a New Model}
To add a new model, users should inherit $\mathsf{BaseModel}$ to design a new $\mathsf{Model}$ class, in which they need to implement the $\mathsf{build\_model()}$ and $\mathsf{forward()}$ functions.
In $\mathsf{build\_model}$, users build the model, initialize the parameters and set up the loss function. While in $\mathsf{forward()}$, users use the model to predict the result or calculate the loss for the input data.
Indeed, users can leverage the encapsulated layers and functions from the utility files to implement the two functions, which are decoupled from existing CRS models and may be useful in most cases.

\ignore{Besides, in some case, one needs to re-implement some functions in the $\mathsf{DataLoader}$ and $\mathsf{System}$ classes.
Specifically, the function $\mathsf{\_batchify}$ in $\mathsf{DataLoader}$ should be modified to match the input batches.
In $\mathsf{System}$ class, users can modify the function $\mathsf{step}$ to adjust the training process of the new model.
It is worth noting that minor modifications are enough to adapt the two functions for integrating the new model.}

\section{Conclusion}
In this paper, we have released a new conversational recommender system (CRS) toolkit called CRSLab, which is the first open-source CRS toolkit for research purpose. 
In CRSLab, we offer a unified and extensible framework with highly-decoupled modules to develop a CRS.
Based on this framework, we integrate comprehensive benchmark datasets and models. So far, we have incorporated 6 commonly-used datasets and implemented 18 models in our toolkit.
Besides, CRSLab also provides extensive automatic evaluation protocols and a human-machine interactive interface to compare and test different CRSs.

With CRSLab toolkit, we expect to help users quickly implement existing CRSs, ease the developing process of new systems, and set up a benchmark framework for the research of CRS. In the future, we will make continuous efforts to add more datasets and models, and will also consider adding more utilities for improving the usage of our toolkit, such as result visualization and algorithm debugging.

\bibliographystyle{acl_natbib}
\bibliography{emnlp-ijcnlp-2019}

\end{document}